\documentclass[conference]{IEEEtran}
\IEEEoverridecommandlockouts
\usepackage{cite}
\usepackage{amsmath,amssymb,amsfonts}
\usepackage{algorithmic}
\usepackage{graphicx}
\usepackage{textcomp}
\usepackage{xcolor}
\def\BibTeX{{\rm B\kern-.05em{\sc i\kern-.025em b}\kern-.08em
    T\kern-.1667em\lower.7ex\hbox{E}\kern-.125emX}}
\begin{document}

\title{Structure-Preserving Patch Decoding for \\ Efficient Neural Video Representation
}

\author{\IEEEauthorblockN{Taiga Hayami}
\IEEEauthorblockA{\textit{Graduate School of FSE,} \\
\textit{Waseda University}\\
Tokyo, Japan \\
hayatai17@fuji.waseda.jp}
\and
\IEEEauthorblockN{Kakeru Koizumi}
\IEEEauthorblockA{\textit{Graduate School of FSE,} \\
\textit{Waseda University}\\
Tokyo, Japan \\
kkeverio@ruri.waseda.jp}
\and
\IEEEauthorblockN{Hiroshi Watanabe}
\IEEEauthorblockA{\textit{Graduate School of FSE,} \\
\textit{Waseda University}\\
Tokyo, Japan \\
hiroshi.watanabe@waseda.jp}
}

\maketitle

\begin{abstract}
Implicit neural representations (INRs) are the subject of extensive research, particularly in their application to modeling complex signals by mapping spatial and temporal coordinates to corresponding values.
When handling videos, mapping compact inputs to entire frames or spatially partitioned patch images is an effective approach.
This strategy better preserves spatial relationships, reduces computational overhead, and improves reconstruction quality compared to coordinate-based mapping.
However, predicting entire frames often limits the reconstruction of high-frequency visual details. 
Additionally, conventional patch-based approaches based on uniform spatial partitioning tend to introduce boundary discontinuities that degrade spatial coherence.
We propose a neural video representation method based on Structure-Preserving Patches (SPPs) to address such limitations.
Our method separates each video frame into patch images of spatially aligned frames through a deterministic pixel-based splitting similar to PixelUnshuffle.
This operation preserves the global spatial structure while allowing patch-level decoding.
We train the decoder to reconstruct these structured patches, enabling a global-to-local decoding strategy that captures the global layout first and refines local details. 
This effectively reduces boundary artifacts and mitigates distortions from naive upsampling.
Experiments on standard video datasets demonstrate that our method achieves higher reconstruction quality and better compression performance than existing INR-based baselines.
\end{abstract}

\begin{IEEEkeywords}
Implicit neural representation, video compression, video representation.
\end{IEEEkeywords}

\section{Introduction}

Implicit Neural Representations (INRs) have gained attention as compact and resolution-independent alternatives to conventional explicit representations for modeling complex signals. 
By parameterizing data as continuous functions of spatial and temporal coordinates using neural networks, INRs significantly reduce memory usage while enabling smooth interpolation and reconstruction of signals at arbitrary resolutions.
These properties have made INRs effective in areas ranging from novel-view synthesis \cite{nerf, mipnerf, dnerf, wild} to static image representation \cite{coin, coinpp, coolchic, c3} and, more recently, to video representation \cite{nerv, enerv, hnerv}.

\begin{figure}[tb]
\centerline{\includegraphics[width=.9\columnwidth]{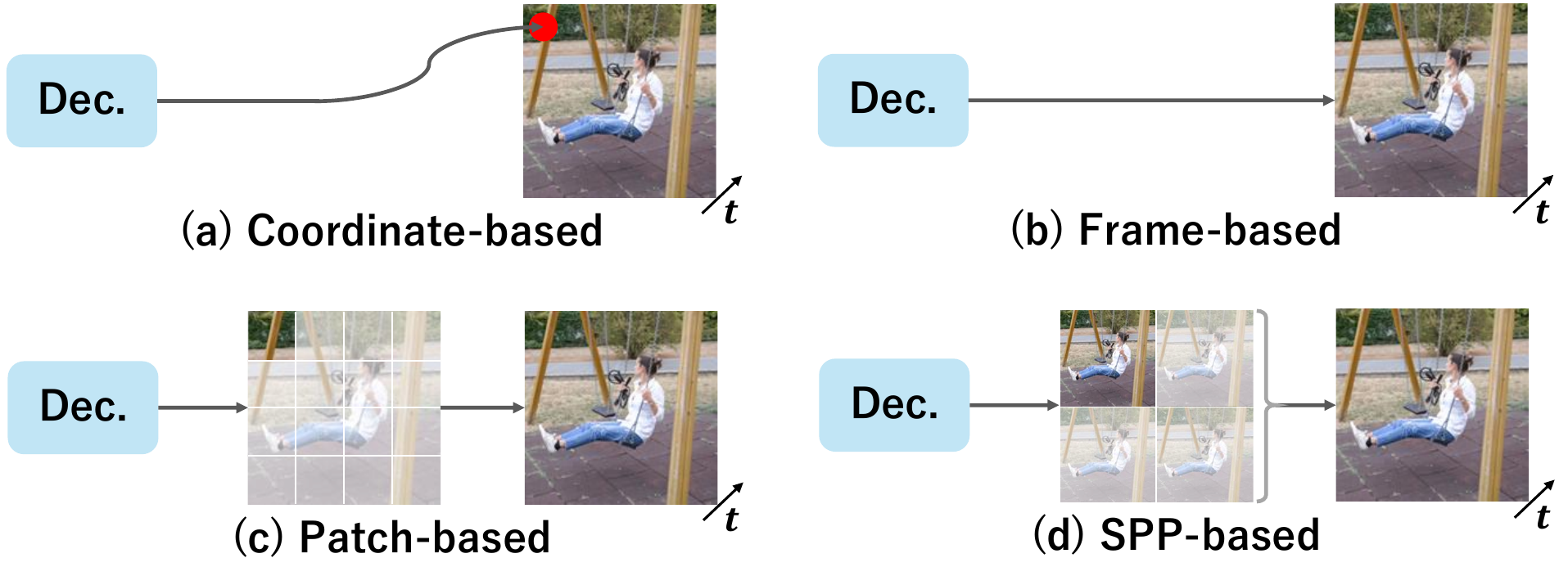}}
\caption{
Comparison of decoding strategies for neural video representation. 
(a) Coordinate-based: predicts pixel color from spatial and temporal coordinates. 
(b) Frame-based: generates entire frames from embeddings. 
(c) Patch-based: reconstructs frames from uniformly divided patches. 
(d) SPP-based (ours): arranges pixels into spatially aligned patches for structure-preserving, global-to-local reconstruction.
}
\label{comp}
\end{figure}

Existing INR-based video representation methods can be categorized into three primary types, depending on their decoding granularity: (1) coordinate-based approaches \cite{siren, bias}, which predict each pixel from its spatiotemporal coordinates; (2) frame-based approaches \cite{nerv, enerv, hnerv}, which generate entire frames from compact input; and (3) patch-based approaches \cite{psnerv, nirvana}, which reconstruct frames by predicting spatially partitioned patches.
While coordinate-based approaches provide fine-grained spatial control and support resolution scalability, their independent pixel-wise processing limits the ability to capture global scene structure and incurs high computational costs.
To address these issues, frame-based methods generate entire frames from compact temporal or content embeddings, enabling faster inference and more effective modeling of global structure.
However, they often struggle to preserve high-frequency components due to upsampling artifacts and the spectral bias of neural networks, which favors low-frequency components \cite{sb1, sb2, sb3, finer}.
In addition, patch-based decoding lies between these extremes. 
Reconstructing each frame from smaller spatial regions combines the local detail fidelity of pixel-based methods with the global structural modeling of frame-based methods.
However, uniform spatial partitioning often leads to discontinuities across patch boundaries, as independently reconstructed regions lack alignment. 
As a result, visual artifacts such as seams and diminished perceptual quality occur.

\begin{figure*}[tb]
\centerline{\includegraphics[width=2.0\columnwidth]{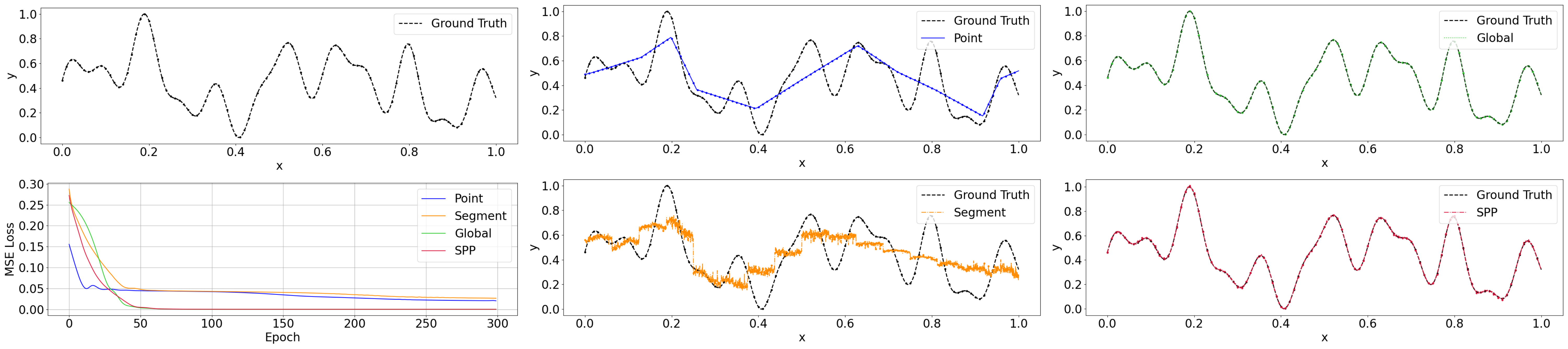}}
\caption{
Toy experiment comparing the fitting performance of one-dimensional signals. 
Top row (left to right): ground-truth signal, point-wise fitting, and global fitting.
Bottom row (left to right): training loss curves, segment-wise fitting, and the proposed Structure-Preserving Patch (SPP)-style fitting.
While conventional segment-wise approaches suffer from boundary discontinuities, the proposed SPP-style approach maintains structural continuity and enables accurate local fitting.
}
\label{toy}
\end{figure*}

In order to overcome these limitations, we propose a Structure-Preserving Patches (SPPs) decoding method for neural video representation. 
Rather than dividing frames into spatially disconnected patches, our method applies a pixel arrangement operation, similar to PixelUnshuffle, to generate patch images that preserve the original spatial layout. 
This pixel-based splitting maintains the relative spatial relationships among patches, enabling the network to first capture global scene structure and later focus on refining local details.
To implement this, the decoder is structured so that early layers are guided by the content embedding and temporal index to model global structure, with later layers incorporating the patch index to specialize in local refinement.
This global-to-local reconstruction strategy reduces patch boundary artifacts and enhances visual coherence.
We validate our approach on multiple benchmark video datasets. 
The experimental results demonstrate that our method achieves superior reconstruction quality and compression performance compared to existing INR-based video representation methods.

\section{Related works}
\subsection{Implicit Neural Representation}
The central concept of INR is to model a signal as a continuous function implemented by a neural network.
Formally, a signal is defined as:
\begin{equation}
y = \Phi_\theta(x),
\end{equation}
where $x$ is a continuous coordinate, $y$ is the signal value at that location, and $\theta$ denotes the learnable parameters of the network.
The network $\Phi_\theta$ is optimized to fit a given target signal.
A seminal work in this field is Neural Radiance Fields (NeRF) \cite{nerf}, which models a 3D scene as a radiance field defined over spatial coordinates and viewing directions. 
NeRF demonstrated that photo-realistic view synthesis is possible by optimizing a fully connected network to predict color and density values along camera rays. 
Although computationally demanding, NeRF highlights the potential of neural functions to model complex continuous structures without relying on explicit mesh or voxel representations.
By leveraging periodic activation functions, SIREN \cite{siren} reduces the spectral bias inherent in ReLU-based networks and improves the network’s ability to capture fine details and textures.
These works show that compact and continuous neural representations have inspired the application of INRs to spatio-temporal domains, including video data. 

\begin{figure*}[tb]
\centerline{\includegraphics[width=1.6\columnwidth]{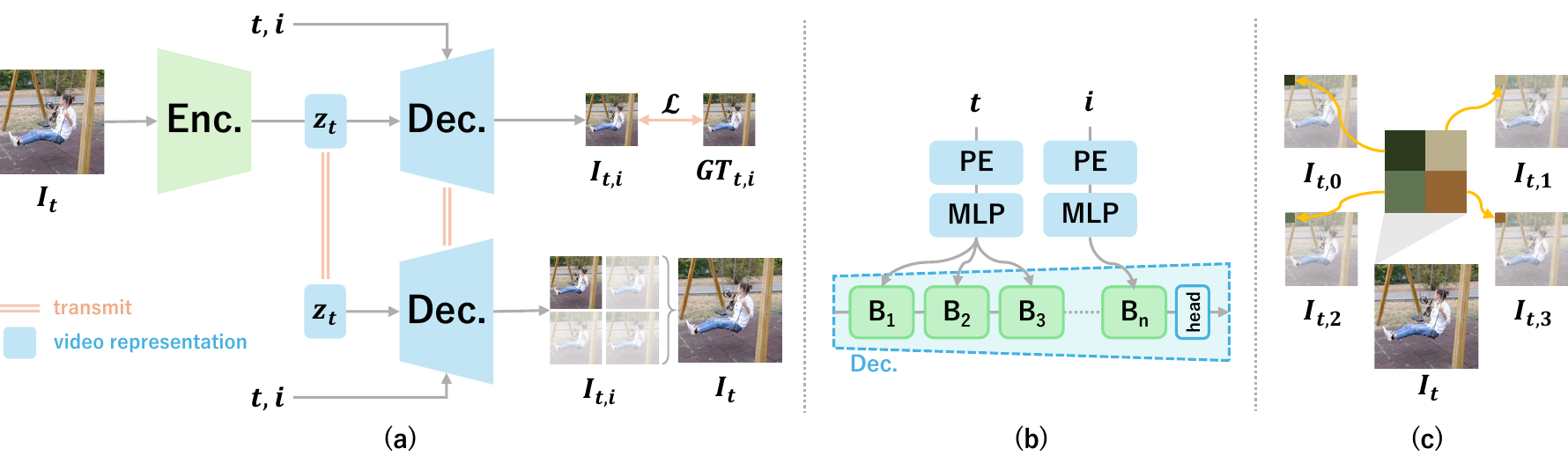}}
\caption{
Overview of the proposed Structure-Preserving Patches (SPPs) based video representation framework.
(a) Overall architecture. 
An input frame $I_t$ is encoded into a frame-level embedding $z_t$, which is then decoded into patch images $I_{t,i}$ using the temporal index $t$ and patch index $i$. 
The full frame is reconstructed by spatially rearranging the decoded patches. 
(b) Decoder details. 
Temporal and patch indices are embedded using positional encodings and MLPs. 
Early decoder layers are conditioned on $t$ to capture global structure, while later layers use $i$ to refine local patch details. 
(c) Patch construction via SPPs. 
Each frame is split into multiple patch images (e.g., $I_{t,0}$, $I_{t,1}$, $I_{t,2}$, $I_{t,3}$) that preserve spatial consistency. 
This design supports a smooth global-to-local reconstruction process.
}
\label{pipeline}
\end{figure*}

\subsection{Neural Video Representation}

INRs have recently been explored as an alternative paradigm for video representation and compression.
A representative example is NeRV \cite{nerv}, which encodes an entire video in the weights of a neural network that maps a frame index to its corresponding RGB frame. 
Unlike conventional INR with coordinate mapping, NeRV treats the video as a global function of learned time, effectively embedding the video content into the network parameters.
Video decoding then requires only a forward pass for a given frame index.
This formulation means that compressing the network (through techniques like weight pruning or quantization) effectively compresses the video. 
Building on this idea, E-NeRV \cite{enerv} improves NeRV by disentangling spatial and temporal components. 
It learns separate embeddings for each domain and combines them during decoding via Adaptive Instance Normalization (AdaIN) \cite{adain}, thereby improving both representational efficiency and compression performance.
HNeRV \cite{hnerv} further builds on this concept by replacing content-agnostic embeddings derived from the frame index with content-adaptive embeddings. 
This adjustment allocates greater modeling capacity to complex or high-frequency frames.
More recently, BoostingNeRV \cite{boost} introduced a conditional decoding framework that can enhance various existing INR-based video models. 
It modulates intermediate features using a temporal affine transformation based on frame indices, enabling improved reconstruction performance across different architectures.

Recent works have also explored alternative decoding strategies to improve reconstruction quality and scalability.
For example, PS-NeRV \cite{psnerv} reconstructs video frames on a patch-wise basis, rather than decoding full frames at once. 
This design provides flexibility in handling high-resolution content and facilitates parallel computation, but often introduces visible artifacts along patch boundaries due to the lack of spatial continuity. 
To better capture video dynamics, DNeRV \cite{dnerv} adopts a dual-stream architecture.
One stream models the static background, while the other captures inter-frame differences, allowing motion to be represented implicitly.
Several other approaches introduce hierarchical model structures \cite{dsnerv, hinerv, pnerv}, incorporating explicit residual connections \cite{resnerv, rcnerv}, or emphasizing the preservation of frequency-domain characteristics \cite{nvrrrcp, snerv, srnerv}, in order to mitigate issues such as temporal flickering or poor reconstruction of fine details.
Our approach follows this research direction, focusing on patch-based representation.
Distinct from prior patch-based methods, we explicitly aim to preserve global spatial structure within the decoding process to mitigate boundary artifacts and improve perceptual consistency.

\section{Proposed method}
\subsection{Overview}
We propose a neural video representation method based on Structure-Preserving Patches (SPPs), as illustrated in Fig. \ref{pipeline}.
In this method, each video frame is decoded from a set of spatially aligned patch images.
To retain the spatial layout of the original frame, we employ a deterministic pixel-based splitting operation, similar to PixelUnshuffle, instead of dividing frames into independently processed tiles as in conventional patch-based methods.
This operation allows the network to exploit redundancy across patch images that share a coherent spatial arrangement, enhancing its ability to model global scene structure.
To further improve reconstruction quality, we design the decoder to follow a global-to-local fitting strategy.
In the early layers, the decoder models global structure using a content embedding and temporal index.
Subsequently, patch indices are incorporated into the later layers to refine local details within each patch.
This hierarchical decoding process enables accurate reconstruction of both global layout and high-frequency visual details.

\subsection{Motivation}
To highlight the importance of the global-to-local strategy, we conduct a simple one-dimensional signal-fitting experiment (Fig. \ref{toy}).  
We compare four strategies:  
(1) \textit{Point-wise fitting}, where the signal is fitted at each coordinate independently;  
(2) \textit{Segment-wise fitting}, where the signal is divided into fixed-length segments, and each segment is fitted independently; 
(3) \textit{Global fitting}, where the entire signal is fitted as a single input; and  
(4) \textit{SPP-style fitting}, our proposed approach, which fits an arranged version of the signal based on fixed position samples, mimicking the structure-preserving patch mechanism.
The results show that the global and SPP-style strategies achieve superior fitting accuracy by preserving the overall structure of the signal.  
In contrast, point-wise and segment-wise methods prioritize local fitting and fail to maintain global consistency.  
These observations suggest that modeling the global structure first, followed by local refinement, is essential for accurate signal reconstruction.
This insight forms the foundation of our global-to-local decoding framework used for video representation.

\begin{figure*}
\centerline{\includegraphics[width=2.0\columnwidth]{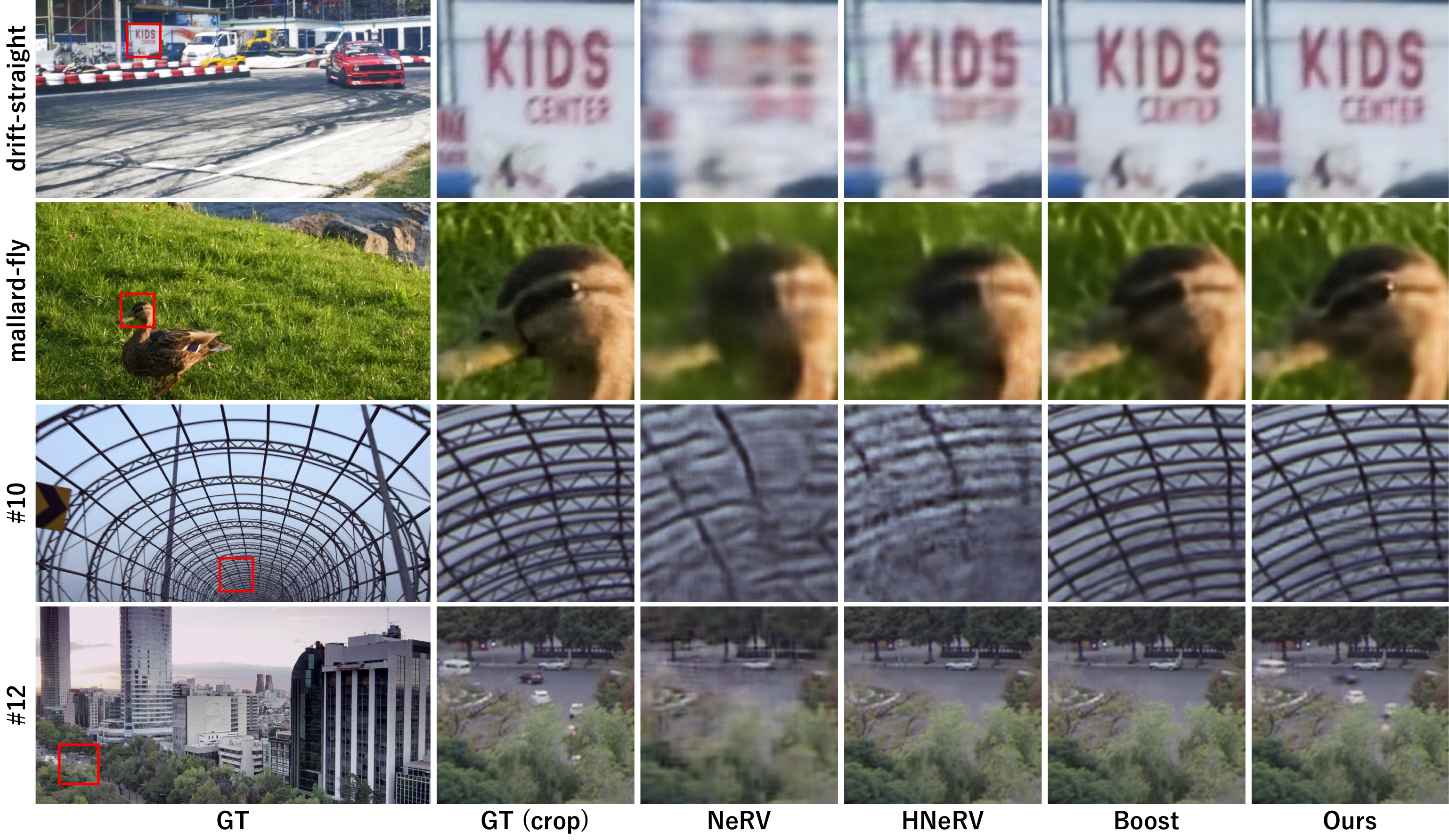}}
\caption{
Visualization of reconstructed frames from different sequences. 
From top to bottom: ``drift-straight'' and ``mallard-fly'' from the DAVIS dataset, and sequences \#10 and \#12 from the MCL-JCV dataset. 
Our Structure-Preserving Patches (SPPs) based method maintains sharp edges and consistent spatial structure across frames, demonstrating effective reconstruction quality across diverse scenes.
}
\label{vis}
\end{figure*}

\subsection{Structure-Preserving Patch Decoding}
Our framework predicts each video frame at the patch level using a neural network that follows a global-to-local decoding strategy, as illustrated in Fig.\ref{pipeline}.  
For each input frame $I_t$ at time $t$, we divide it conceptually into $P$ spatially structured patches.  
To implement this, we apply a pixel arrangement operation, similar to PixelUnshuffle, which transforms the frame into $P$ patch images: $I_{t, 0}, I_{t, 1}, \dots, I_{t, P-1}$.  
The encoder, implemented using ConvNeXt blocks~\cite{convnext, hnerv}, first processes the full frame $I_t$ to produce a compact frame-level embedding $z_t$.  
This embedding is passed to a decoder, which generates each patch image $I_{t, i}$ conditioned on two indices: the temporal index $t$ and the patch index $i$.  
In the decoder, early layers are modulated only by $t$, allowing the model to capture the global structure shared across all patches in the frame.  
Later layers are additionally conditioned on $i$, enabling the network to refine local details specific to each patch.
Once all patch outputs $I_{t, i}$ are obtained, we reconstruct the full-resolution frame $\hat{I}_t$ by spatially rearranging them.  
This hierarchical conditioning mechanism allows the decoder to first build a coherent global structure, then specialize to local content, resulting in improved reconstruction quality.  
In addition, the structure-preserving nature of our patch generation helps avoid boundary discontinuities typically associated with naive patch-based decoding.
The decoder is composed of multiple blocks $B_n$, each of which integrates the SNeRV block and the TAT Residual Block introduced in BoostingNeRV \cite{boost}.

\begin{table}[t]
\caption{Comparison of reconstructed video quality \\ on DAVIS dataset ($1.5\text{M}$, $640\times1280$)}
\begin{center}
\begin{tabular}{c|ccc}
\hline
Method   & PSNR $\uparrow$   & MS-SSIM $\uparrow$ & LPIPS $\downarrow$ \\ \hline\hline
NeRV     & 28.60             & 0.8811             & 0.4150             \\ 
HNeRV    & 30.69             & 0.9146             & 0.3476             \\ 
Boost    & \underline{33.53} & \underline{0.9604} & \underline{0.2674} \\
Ours     & \textbf{34.23}    & \textbf{0.9643}    & \textbf{0.2539}    \\ \hline
\end{tabular}
\end{center}
\label{davis}
\end{table}

\subsection{Loss Function}
For each patch $i$ of a frame, we define a composite loss $\mathcal{L}_i$ that accounts for spatial accuracy, perceptual quality, and frequency-domain consistency:
\begin{equation}
\mathcal{L}_i = w_i(\alpha \mathcal{L}_1(x_i, \hat x_i) + \beta\mathcal{L}_{\text{MS-SSIM}}(x_i, \hat x_i)) + \mathcal{L}_{\text{freq}}(x_i, \hat x_i),
\end{equation}
\begin{equation}
\mathcal{L}_{\text{freq}}(x, \hat x) = \mathcal{L}_1(\text{FFT}(x), \text{FFT}(\hat x)),
\end{equation}
where $x_i$ and $\hat{x}_i$ denote the ground-truth and predicted patch images, respectively.  
The term $\mathcal{L}_{1}$ is the standard L1 pixel loss, and $\mathcal{L}_{\text{MS-SSIM}}$ is a multi-scale structural similarity loss, which promotes perceptual quality. 
The coefficients $\alpha$ and $\beta$ balance their relative contributions.  
In addition, The frequency domain term $\mathcal{L}_{\text{freq}}$ is computed by applying the Fast Fourier Transform (FFT) to both images and measuring their L1 difference.

In our SPP-based decoding scheme, patch-specific variations are adjusted only at the later layers of the decoder.  
This localized modulation makes it difficult for the model to learn patches that differ significantly from their neighbors. 
To address this, we introduce an adaptive patch weighting factor $w_i$ as follows:  
\begin{equation}
w_i = \frac{\sum_{j \ne i} \| x_i - x_j \|_1}{\sum_{k} \sum_{j \ne k} \| x_k - x_j \|_1 + \epsilon},
\end{equation}
where $\epsilon$ is a small constant to avoid division by zero.
This formulation normalizes the weights across all patches.  
The model focuses more on learning consistent structures while maintaining flexibility for difficult regions.

\begin{table}[t]
\caption{Comparison of reconstructed video quality \\ on MCL-JCV dataset ($1.5\text{M}$, $640\times1280$)}
\begin{center}
\begin{tabular}{c|ccc}
\hline
Method   & PSNR $\uparrow$   & MS-SSIM $\uparrow$ & LPIPS $\downarrow$ \\ \hline\hline
NeRV     & 31.64             & 0.9217             & 0.4126             \\ 
HNeRV    & 33.47             & 0.9417             & 0.3571             \\ 
Boost    & \underline{35.60} & \underline{0.9638} & \underline{0.3039} \\
Ours     & \textbf{35.94}    & \textbf{0.9654}    & \textbf{0.2936}    \\ \hline
\end{tabular}
\end{center}
\label{mcl}
\end{table}

\begin{table*}[t]
\caption{Comparison of reconstructed video quality on UVG dataset in PSNR/MS-SSIM ($3\text{M}$, $1080\times1920$)}
\begin{center}
\begin{tabular}{c|ccccccc|c}
\hline
Method   & Beauty         & Bosphorus      & HoneyBee       & Jockey         & ReadySetGo     & ShakeNDry      & YachtRide      & Average \\ \hline\hline
NeRV     & 32.93 / 0.8843 & 33.09 / 0.9288 & 37.08 / 0.9780 & 31.06 / 0.8767 & 24.66 / 0.8205 & 32.67 / 0.9264 & 27.75 / 0.8573 & 31.22 / 0.8937\\
HNeRV    & 33.18 / 0.8876 & 17.57 / 0.5885 & 38.97 / 0.9838 & 31.75 / 0.8884 & 25.09 / 0.8358 & 33.73 / 0.9337 & 27.95 / 0.8546 & 29.44 / 0.8470\\
Boost    & \underline{33.70} / \underline{0.8980} & \underline{35.81} / \underline{0.9631} & \underline{39.52} / \underline{0.9852} & \underline{33.84} / \textbf{0.9253}    & \underline{27.72} / \textbf{0.9070}    & \underline{35.55} / \underline{0.9532} & \underline{29.03} / \textbf{0.8967}    & \underline{33.45} / \underline{0.9311}\\
Ours     & \textbf{33.82}    / \textbf{0.8992}    & \textbf{36.08}    / \textbf{0.9644}    & \textbf{39.60}    / \textbf{0.9854}    & \textbf{34.21}    / \underline{0.9249} & \textbf{28.03}    / \underline{0.9054} & \textbf{35.96}    / \textbf{0.9584}    & \textbf{29.33}    / \underline{0.8945} & \textbf{33.70}    / \textbf{0.9312}   \\ \hline
\end{tabular}
\end{center}
\label{uvg}
\end{table*}

\section{Experiment}

\begin{table}[tb]
\caption{PSNR over training epochs on the Bunny video \\ ($1.5\text{M}$, $640\times1280$)}
\begin{center}
\begin{tabular}{c|cccccc}
\hline
Epoch    & 50                & 100               & 150               & 200               & 250               & 300               \\ \hline\hline
NeRV     & 26.95             & 29.40             & 30.43             & 30.96             & 31.14             & 31.33             \\ 
HNeRV    & 29.93             & 33.32             & 34.86             & 35.61             & 36.06             & 36.35             \\ 
Boost    & \underline{34.90} & \underline{37.35} & \underline{38.12} & \underline{38.62} & \underline{38.70} & \underline{39.03} \\ 
Ours     & \textbf{37.80}    & \textbf{38.62}    & \textbf{39.00}    & \textbf{39.18}    & \textbf{39.31}    & \textbf{39.40}    \\ \hline
\end{tabular}
\end{center}
\label{bunny}
\end{table}

We evaluate the effectiveness of the proposed SPP-based video representation method across four benchmark datasets: Bunny \cite{bunny}, DAVIS \cite{davis}, MCL-JCV \cite{mcl}, and UVG \cite{uvg}. 
The Bunny dataset consists of a single sequence with a resolution of $720\times1280$ and 132 frames. 
To ensure stable training, the frames are cropped to $640\times1280$. 
This adjustment is made because HNeRV failed to converge when trained on high-resolution input, as observed in the ``Bosphorus'' sequence in Table \ref{uvg}.
The DAVIS dataset consists of 50 natural video sequences with resolution $1080\times1920$ (cropped $640\times1280$). 
Each sequence is relatively short (25 to 104 frames). 
Similarly, MCL-JCV contains 30 sequences at $1080\times1920$ (cropped to $640\times1280$), with lengths ranging from 120 to 150 frames. 
The UVG dataset comprises 7 long sequences at full HD resolution ($1080\times1920$), each containing 300 or 600 frames.
Our method is compared with existing implicit neural video representation baselines, including NeRV, HNeRV, and Boosting-HNeRV (``Boost'').

We adjust the stride configuration in the decoder to match the spatial resolution: the stride list is set to [5, 4, 2, 2, 2] for $640\times1280$, and [5, 3, 2, 2] for $1080\times1920$.
The number of patches is fixed at $P=4$, and we empirically set the loss weighting coefficients to $\alpha=42$ and $\beta=18$.
The default model size is 1.5M parameters, except for the UVG dataset at $1080\times1920$ resolution, where a 3.0M parameter model is used.
For quantitative evaluation, we use PSNR, MS-SSIM, and LPIPS.
In the video compression setting, we use the UVG dataset with a resolution of $1080\times1920$ and subsample each sequence to 60 or 120 frames. 
Each video is encoded into multiple models of varying sizes by adjusting the network width. 
We apply quantization and entropy coding to these models, enabling rate-distortion (RD) analysis.

\begin{figure}[t]
\centerline{\includegraphics[width=1.0\columnwidth]{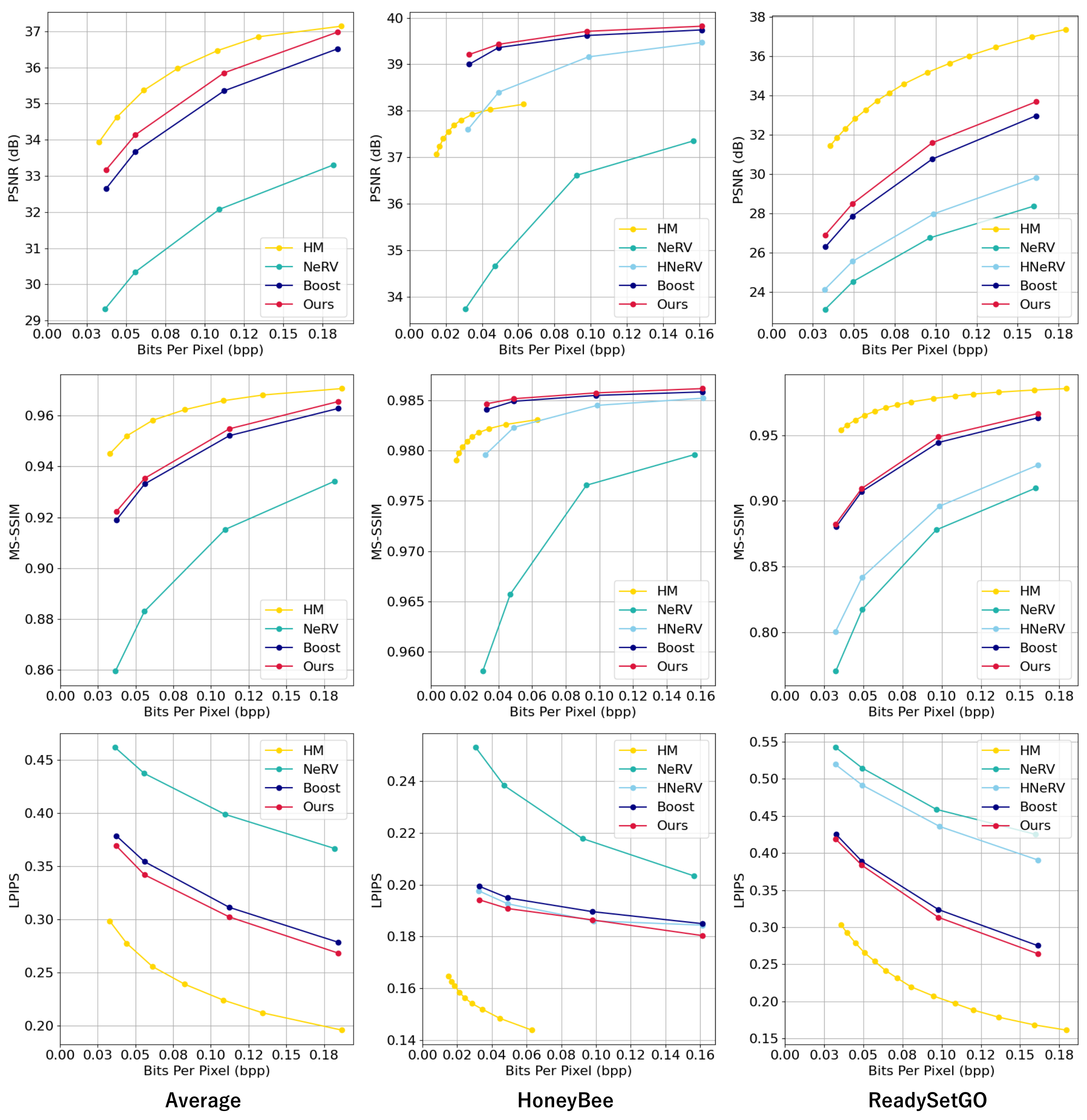}}
\caption{Compression result on UVG dataset.}
\label{rd}
\end{figure}

Quantitative results on DAVIS and MCL-JCV are summarized in Tables \ref{davis} and \ref{mcl}, respectively. 
These tables show the average reconstruction quality across each dataset.
Table \ref{uvg} reports PSNR and MS-SSIM scores for each sequence in the UVG dataset. 
Our method consistently achieves higher scores in most sequences, demonstrating superior reconstruction quality compared to baseline methods.
In addition, Table~\ref{bunny} presents PSNR values at each training epoch for the Bunny dataset, demonstrating not only improved reconstruction performance but also faster convergence compared to competing methods.
Qualitative comparisons are shown in Fig.~\ref{vis}, which visualizes representative frames from DAVIS and MCL-JCV. 
Our method better preserves sharp edges and spatial coherence across a variety of scenes.
Finally, Fig. \ref{rd} presents the rate-distortion curves obtained from UVG. 
Note that HNeRV is excluded from the average curve due to repeated convergence failures during training.
The proposed method demonstrates superior compression performance relative to baseline INR methods. 
However, it does not achieve the same performance as traditional video compression technology, HEVC \cite{hevc}, and there is still potential for improvement in model compression.

\section{Conclusion}
We propose a neural video representation method based on Structure-Preserving Patches (SPPs). 
In our approach, each video frame is split into spatially consistent patch images that maintain the original spatial structure.
This strategy mitigates the degradation commonly seen in conventional methods and contribute to the fitting of global structures.
To facilitate smooth reconstruction, we designed a global-to-local decoder architecture. 
Early decoding layers are conditioned on the temporal index to model the global structure, while later layers incorporate patch-specific information to enhance local details. 
This separation allows the network to balance structural coherence and fine-grained fidelity effectively.
Experimental results demonstrate that our method achieves higher reconstruction quality than existing NeRV-based baselines. 
These findings highlight the potential of structure-aware patch design for improving the fidelity and efficiency of neural video representations.
In future work, we plan to explore adaptive, content-aware patch layouts that dynamically adjust to scene complexity, aiming to further improve spatial coherence and compression performance.

\vspace{12pt}
\end{document}